\documentclass[11pt,a4paper]{article}
\usepackage[hyperref]{emnlp-ijcnlp-2019}
\usepackage{times}
\usepackage{latexsym}

\usepackage{url}

\usepackage{amssymb}
\usepackage{amsmath}
\usepackage{latexsym}
\usepackage{url}
\usepackage{mathtools}
\usepackage{multirow}
\usepackage{xspace}
\usepackage{comment}
\usepackage{flowchart}
\usetikzlibrary{arrows}
\usepackage{booktabs}
\usepackage{subcaption}
\usepackage{hyperref}
\usepackage{enumitem}
\usepackage{multirow}
\usepackage[algo2e,vlined,ruled]{algorithm2e}
\usepackage{tikz}
\usepackage{pgfplots}
\usepackage{enumitem}
\usepackage{todonotes}
\usepackage{enumitem}
\usepackage[utf8]{inputenc}
\usepackage{CJKutf8}

\DeclareMathAlphabet{\pazocal}{OMS}{zplm}{m}{n}
\DeclareMathAlphabet{\pazocal}{OMS}{zplm}{m}{n}

\aclfinalcopy 

\title{Evaluating Pronominal Anaphora in Machine Translation:\\ An Evaluation Measure and a Test Suite}

\author{Prathyusha Jwalapuram$^*$, Shafiq Joty$^*$$^\S$, Irina Temnikova$^\dagger$, and Preslav Nakov$^\ddagger$\\
  $^*$Nanyang Technological University, Singapore \\
  $^\S$Salesforce Research Asia, Singapore \\
  $^\dagger$Sofia University, Bulgaria\\
  $^\ddagger$Qatar Computing Research Institute, Doha, Qatar\\
  $^*${\tt\{jwal0001,srjoty\}@ntu.edu.sg} \\
  $^\dagger$\tt{irina.temnikova@gmail.com} \\
  $^\ddagger${\tt pnakov@hbku.edu.qa} \\
\\}

\newcommand{\ie}{{\em i.e.,}\xspace}

\newcommand{\resp}{\emph{resp.}\xspace}

\newcommand{\Ls}{\mathcal{L}}

\newcommand{\Ds}{\pazocal{D}}

\newcommand{\Ss}{\pazocal{S}}

\newcommand{\real}{\mathbb{R}}
\date{}

\begin{document}

\maketitle
\begin{abstract}
The ongoing neural revolution in machine translation has made it easier to model larger contexts beyond the sentence-level, which can potentially help resolve some discourse-level ambiguities such as pronominal anaphora, thus enabling better translations. Unfortunately, even when the resulting improvements are seen as substantial by humans, they remain virtually unnoticed by traditional automatic evaluation measures like BLEU, as only a few words end up being affected. Thus, specialized evaluation measures are needed. With this aim in mind, we contribute an extensive, targeted dataset that can be used as a test suite for pronoun translation, covering multiple source languages and different pronoun errors drawn from real system translations, for English. We further propose an evaluation measure to differentiate good and bad pronoun translations. We also conduct a user study to report correlations with human judgments.
\end{abstract}

\section{Introduction} 
\label{sec:intro}

Traditionally, machine translation (MT) has been performed at the level of individual sentences, i.e.,~in isolation from the rest of the document. 
This was due to the nature of the underlying frameworks: word-based \cite{J93-2003}, then phrase-based \cite{Koehn:2003:SPT:1073445.1073462}, syntactic \cite{galley-EtAl:2004:HLTNAACL}, and hierarchical \cite{Chiang:2005:HPM:1219840.1219873}. While there have been attempts to model the context beyond the sentence level, e.g., looking at neighboring sentences \cite{D07-1007,P07-1005} or even at the entire document \cite{Hardmeier:2012:DDP:2390948.2391081}, these approaches were still limited by the underlying framework, which was focusing on very narrow contexts.

Then, along came the neural revolution, and the situation changed. Thanks to the attention mechanism, neural translation models such as sequence-to-sequence \cite{DBLP:journals/corr/BahdanauCB14} and the Transformer \cite{Vaswani-17-transformer} could model much broader context. While initially translation was still done in a sentence-by-sentence fashion, researchers soon realized that going beyond the sentence level has become easier and more natural than before and recent work has successfully exploited this \cite{Bawden-NAACL18,Voita-ACL18}. 
This is an exciting research direction as it can help address inter-sentential phenomena such as anaphora, gender agreement, lexical consistency, and text coherence, to mention just a few. 

Unfortunately, going beyond the sentence level typically yields very few changes in the translation output, and even when these changes are seen as substantial by humans, they remain virtually unnoticed by typical MT evaluation measures such as BLEU \cite{papineni2002bleu}, which are known to be notoriously problematic for the evaluation of discourse-level aspects in MT \cite{Hardmeier-14}. 

The limitations of BLEU are well-known and have been discussed in detail in a recent study \cite{reiter2018structured}. It has long been argued that as the quality of machine translation improves, there will be a singularity moment when existing evaluation measures would be unable to tell whether a given output was produced by a human or by a machine. Indeed, there have been recent claims that human parity has already been achieved \cite{DBLP:journals/corr/abs-1803-05567}, but it has also been shown that it is easy to tell apart a human translation from a machine output when going beyond the sentence level \cite{Samuel-EMNLP-18}.  
Overall, it is clear that there is a need for machine translation evaluation measures that look beyond the sentence level, and thus can better appreciate the improvements that a discourse-aware MT system could potentially bring.

Alternatively, one could use diagnostic test sets that are designed to evaluate how a target MT system handles specific discourse phenomena \cite{Bawden-NAACL18,rios-mller-sennrich:2018:WMT}. There have also been proposals to use semi-automatic measures and test suites instead of fully automatic evaluation measures \cite{Liane-EMNLP18}. 

Here we propose a targeted dataset for MT evaluation with a focus on anaphora and a specialized evaluation measure trained on this dataset. The measure performs pairwise evaluations: it learns to distinguish good vs. bad translations of pronouns, without being given specific signals of the errors. It has been argued that pairwise evaluation is useful and sufficient for machine translation evaluation \cite{Guzman2015a, Guzmn2017MachineTE}. In particular, \citet{Duh2008} has shown that ranking-based evaluation measures can achieve higher correlations with human judgments, as rankings are simpler to obtain from humans and to train models on, while also directly achieving the purpose of comparing two systems.

{Note that while it may be possible to rank translations using strong pre-trained conditional language models {like GPT \cite{GPT-1},} all kinds of errors would influence the score - it would not be targeted towards a specific source of error, such as anaphora here. Our model provides a way to do this, and we demonstrate that our model indeed focuses on pronouns.}

Although the pronoun test suite naturally consists of the source text paired with the reference translation, our pronoun evaluation measure is independent of the source language. Moreover, we use real MT output, which may contain various types of errors.
Our contributions are as follows:

\begin{itemize}
\item We create a dataset for pronoun translation covering multiple source languages and various target English pronouns.
\item We propose a novel evaluation measure that differentiates good pronoun translations from bad ones irrespective of the source language they were translated from.
\item Unlike previous work, both the dataset and the model are based on actual system outputs.

\item Our evaluation measure achieves high agreement with human judgments.
\end{itemize}

We make both the dataset and the evaluation measure publicly available at \href{https://ntunlpsg.github.io/project/discomt/eval-anaphora/}{https://ntunlpsg.github.io/project/discomt/eval-anaphora/}.

\section{Related Work}
\label{sec:rel-work}

Previous work on discourse-aware machine translation and MT evaluation has targeted a number of phenomena such as anaphora, gender agreement, lexical consistency, and coherence.
In this work, we focus on pronoun translation.

Pronoun translation has been the target of a shared task at the DiscoMT and WMT workshops in 2015-2017 \cite{W15-2501,Guillou-16-WMT,Sharid-W17}.
However, the focus was on cross-lingual pronoun prediction, which required choosing the correct pronouns in the context of an existing translation, i.e.,~this was not a realistic translation task. The 2015 edition of the task also featured a pronoun-focused translation task,  which was like a normal MT task except that the evaluation focused on the pronouns only, and was performed manually. In contrast, we have a real MT evaluation setup, and we develop and use a fully automatic evaluation measure.

 More recently, there has been a move towards using specialized test suites
specifically designed to assess system quality for some fine-grained problematic categories, including pronoun translation. For example, the PROTEST test suite \cite{GUILLOU16.327} comprises 250 pronoun tokens, used in a semi-automatic evaluation: the pronouns in the MT output and reference are compared automatically, but in case of no matches, manual evaluation is required. Moreover, no final aggregate score over all pronouns is produced. In contrast, we have a much larger test suite with a fully automatic evaluation measure.

Another semi-automatic system is described in \citet{guillou2018pronoun}. It focused on just two pronouns, \emph{it} and \emph{they}, and was applied to a single language pair. In contrast, we have a fully automated evaluation measure, handle many English pronouns, and cover multiple source languages.

\newcite{Bawden-NAACL18} presented hand-crafted discourse test sets, designed
to test the model's ability to exploit previous source and target sentences, based on 200 contrastive pairs of sentences, where one has a correct and one has a wrong pronoun translation. This alleviates the need for an automatic evaluation measure as one can just count how many times the MT system has generated a translation containing the correct pronoun. 
In contrast, we work with natural texts from pre-existing MT evaluation datasets, we do not require them to be in contrastive pairs, and we have a fully automated evaluation measure. Moreover, we use much larger-scale evaluation datasets.
\newcite{muller-EtAl:2018:WMT} also used contrastive translation pairs, mined from a parallel corpus using automatic coreference-based mining of context, thus minimizing the risk of producing wrong contrastive examples that are both valid translations. Yet, they did not propose an evaluation measure.

Finally, there have been pronoun-focused automatic MT evaluation measures; 
\citet{Liane-EMNLP18} mention only two main ones: APT \cite{APT} and AutoPRF \cite{Hardmeier2010ModellingPA}. Both measures require alignments between the source, reference and system output texts for evaluating the pronoun translations. But automatic alignments are noisy; \citet{Liane-EMNLP18} show that improvements using heuristics are not statistically significant. They also find low agreement between these measures and human judgments, primarily due to the possibility of many translation choices per pronoun.
APT also uses a predetermined list of `equivalent pronouns', obtained for specific pronouns based on a French grammar book and verified through probability counts. This list is used to weight pronouns that are not exact matches, and the accuracy of the pronoun translations is calculated accordingly. \citet{APT} collect such a list for English-French for the pronouns \textit{it} and \textit{they}. This limits the evaluation measure both by language and the pronouns it is applicable to.

In contrast, our framework requires only two candidate translations of the same text as input for comparison; this could be a reference vs. a system translation, or a comparison between two candidate translations (see Section \ref{subsec:further_analysis}).

\section{Dataset Generation}
\label{sec:dataset}

We automatically generate the dataset used to build a pronoun test suite and train our pronoun evaluation model. To avoid generating synthetic data that may not necessarily represent a difficult context (for an MT system to correctly translate the pronouns), we use data from actual system outputs submitted for the WMT translation tasks in 2011--2015 and 2017 \cite{WMT2011,WMT2012,WMT2013,WMT2014,WMT2015,WMT2017}. Using such data means that what is essentially a conditional language model solution, such as the one used by \newcite{Bawden-NAACL18}, has already failed on these examples. 
In particular, we aligned the system outputs with the reference translation using an automatic alignment tool \cite{dyer2013fastalign}, and found examples in which the pronouns did not match the reference translation. This process produces potentially noisy data, as the alignments are automatic and thus not always perfect.

\begin{figure}

{\textbf{Original French input}: \emph{\textbf{Il} était créatif, généreux, drôle, affectueux et talentueux, et il va beaucoup \textbf{me} manquer.} \\[-0.9em]

\textbf{Reference translation}: \emph{\textbf{He} was creative, generous, funny, loving and talented, and \textbf{I} will miss him dearly.}\\[-0.9em]

\textbf{MT system translation}: \emph{\textbf{It} was creative, generous, funny, affectionate and talented, and \textbf{we} will greatly miss.}\\[-0.9em]

\textbf{Generated noisy example 1}: \emph{\textbf{It} was creative, generous, funny, loving and talented, and I will miss him dearly.} \\[-0.9em]

\textbf{Generated noisy example 2}: \emph{He was creative, generous, funny, loving and talented, and \textbf{we} will miss him dearly.}}
   
\caption{Noisy examples generated by substituting an MT-generated pronoun in the reference translation.}

\label{fig:noisy_data_example2}
\end{figure}

\subsection{User Study}

To ensure that the mismatched pronouns are not equally good translations in the given context, we conducted a user study on a subset of the generated data. To focus the study on pronouns and remove the influence that other errors in the MT output may have on the study participants, we generated a noisy candidate by replacing the correct pronoun in the reference translation with the aligned (potentially) incorrect pronoun from the system output. We did this for each differing pronoun in the MT output, so that the difference between the reference and the noisy version is one pronoun only (see Figure~\ref{fig:noisy_data_example2}).

\begin{figure*}[t!]
\centering
	\includegraphics[scale=0.28]{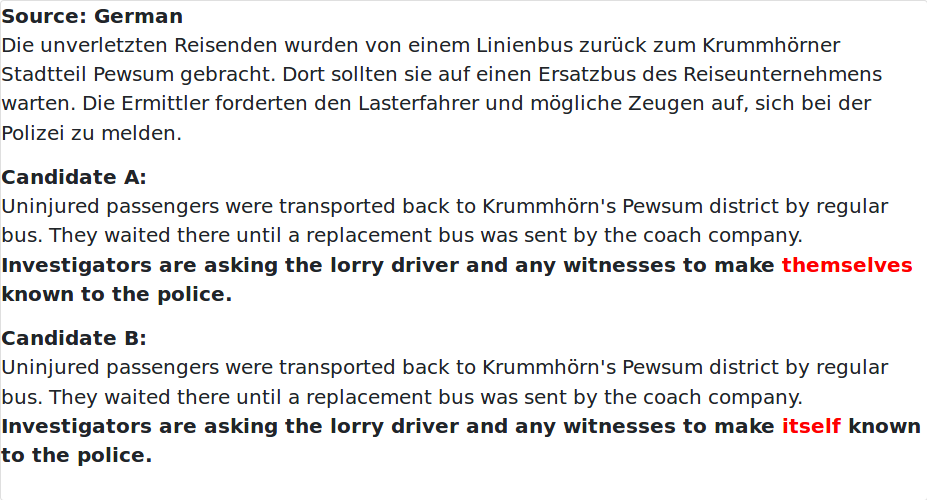}
    \includegraphics[scale=0.28]{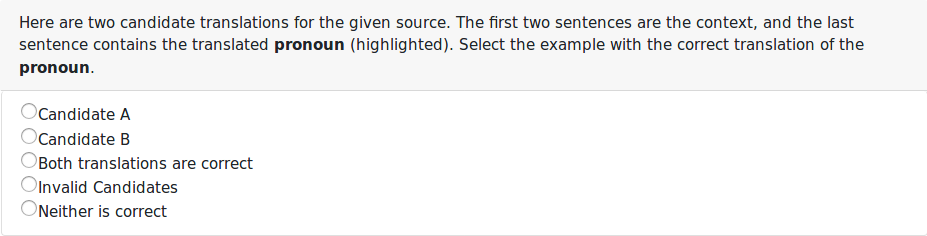}
 
	\caption{Our annotation framework.} \label{fig:study_screenshot}
\end{figure*}

Our goal is to find pronoun pairs (e.g., \emph{He}-\emph{it} in Figure~\ref{fig:noisy_data_example2}) where there is high agreement that the reference is the correct translation, so that we can automatically classify it as a positive example and the MT output as negative. The study participants were fluent in English and were native speakers of Chinese, Russian, French, or German. They were shown the source and two candidate translations (the reference and its noisy version) in random order. The relevant sentence was shown in bold, with the pronoun highlighted. Two previous sentences were given as context; see Figure~\ref{fig:study_screenshot}.

We asked the participants to choose the text with the better pronoun, their choices being candidate A, candidate B, equivalent translations (tie), ``neither is correct'', and ``invalid candidates'' (the highlighted words are not pronouns or are the wrong pronouns due to misalignment).
All examples marked as these last two were excluded from further consideration. Each participant annotated a total of \textbf{500 examples} per language pair. Statistics\footnote{Due to the nature of the dataset, the human annotators are always more likely to choose the reference as the better candidate, which yields a skewed distribution of the annotations; traditional correlation measures such as Cohen's kappa are not robust to this, and thus we report the more appropriate Gwet's AC1/gamma coefficient.\cite{gwetac1}.} are given in Table~\ref{table:agreements}. We also report the proportion of cases where the participants preferred the reference translation over the noisy version (see \textbf{Avg\%Ref}).\footnote{High agreement could also mean that the participants consistently pick the noisy version as the better choice.} We can see that there is high agreement for all language pairs, ranging from 0.82 to 0.89. The ties seem to be the major source of disagreement: excluding them yields agreements in the range of 0.91--0.97.

\begin{table}[t!]
\scalebox{0.72}{\begin{tabular}{r|cccc}

\textbf{Language} & \textbf{Nb. of} & \textbf{Avg\%} & \textbf{AC1} & \textbf{AC1} \\
\textbf{Pair} & \textbf{Ann.} & \textbf{Ref} & \textbf{(Incl. Ties)} & \textbf{(Excl. Ties)} \\
\midrule

Russian$\rightarrow$English & 3 & 80.2 & 0.82 & 0.92 \\

French$\rightarrow$English & 2 & 83.9 & 0.86 & 0.96\\

German$\rightarrow$English & 2 & 84.3 & 0.89 & 0.97\\
Chinese$\rightarrow$English & 3 & 86.0 & 0.86 & 0.91\\ 
\quad- - Only English & 3 & 85.3 & 0.84 & 0.92 \\
\bottomrule
\end{tabular}}

\caption{Inter-annotator agreement.}
\label{table:agreements}

\end{table}

In order to measure the effect, if any, that the source text has on annotator's choices, we also conducted a study without the source text using the texts from the Chinese$\rightarrow$English study. Participants were only shown the English texts: the reference vs. the noisy sentence, with the context as before. 
We see that the agreement for the English-only setup is also fairly high (0.84); the overall agreement between all 6 participants (3 from Chinese$\rightarrow$English and 3 from only English) is 0.85. We observe very similar agreement of 0.91 (Chinese$\rightarrow$English) and 0.92 (Only English) when ties are excluded, with the overall agreement being 0.90 between the 6 participants. However, further analysis showed that although both groups disagreed on about 10\% of the samples, only 2\% of the samples were common to both groups, showing that the sources of disagreement between the two groups are different. Possibly having the source context helps disambiguate the other 8\% of the cases, while also introducing ambiguity that does not seem to be an issue for the participants who saw only English texts. See Figure \ref{fig:noisy_data_example2} for an example where the source is helpful; noisy example 2 would be acceptable, except that the original French text uses a singular pronoun. 
\begin{figure}

\begin{CJK*}{UTF8}{gbsn}
\textbf{Source Text}: 不过现在，她只想享受当下。 我不想说这是我的最后一场比赛。 这会给我带来太大的压力。
\clearpage\end{CJK*}
\vspace{-0.6em}
\textbf{Reference translation}: \emph{For now she just wants to enjoy the moment. I didn't want to say \textbf{this} was my last race. That would have meant too much pressure.}\\[0.5em]
\textbf{Noisy candidate}: \emph{For now she just wants to enjoy the moment. I didn't want to say \textbf{that} was my last race. That would have meant too much pressure.}
   
\caption{Low-agreement example: Chinese-English.}

\label{fig:filtered_example}
\end{figure}
\normalsize

However, the disagreements form a small part of the dataset; we also filter out all pronoun pairs with low agreements from further use. See Figure \ref{fig:filtered_example} for a low-agreement example.

\subsection{Pronoun Test Suite for MT Systems}

The source sentences can also be used as a \textbf{test suite} for MT systems to check their pronoun translations:
it can be considered a challenging, diagnostic test set for pronoun translation, covering a range of errors like gender (he/she$\rightarrow$it), number (they$\rightarrow$it), animacy (who$\rightarrow$which), syntactic role (e.g.~subject/object: he$\rightarrow$him), and others; see Appendix for a complete list.

\noindent WMT test sets come from news articles; the context is available, so the test suite is particularly suitable for discourse-level MT systems. Data is available for each source language for which English translations are generated in WMT tasks: German, Czech, French, and others (Table~\ref{table:pronountestsuite}).

The corresponding noisy versions of the reference are also generated, although there is some noise in this dataset. However, the subset used for the study is curated in some sense, since human judgments are available. This data can serve as a more refined test suite: not only useful for checking agreements with human judgments, but also identifying equivalent pronoun translations in context, as the data is also annotated for ties.

\begin{table}[t!]
\scalebox{0.80}{\begin{tabular}{c|l|c}

& \textbf{Test Data} & \textbf{Unique} \\
\textbf{Source Language} & \textbf{from WMT Years} & \textbf{Source Contexts} \\

\midrule
German & 2011-2015,17 & 7,823 \\
Czech & 2011-2015,2017 & 6,713 \\ 
French& 2011-2015 & 4,659 \\
Russian & 2013,2014,2017 & 4,513 \\ 
Spanish & 2011-2013 & 4,417 \\
Finnish & 2015,2017 & 1,551 \\
Turkish & 2017 & 1,372 \\
Hindi& 2014 & 921 \\
Chinese & 2017 & 696 \\
Latvian & 2017 & 652 \\

\bottomrule
\end{tabular}}

\caption{Pronoun test suite for MT systems: English as a target, and various languages as a source.}
\label{table:pronountestsuite}

\end{table}

\section{The Evaluation Measure}
\label{sec:model}

While \emph{diagnostic datasets} allow us to evaluate MT systems with respect to specific discourse-level phenomena, an automatic discourse-aware \emph{evaluation measure} is useful not only for evaluation but also for tuning MT systems. Moreover, an evaluation measure that only looks at the target language (which is {computationally} feasible, even if not ideal, as our study above has shown) offers  additional benefits; we can train it for a specific target language without requiring a separate dataset for each source-target language pair. Below, we propose such a measure for pronoun translation. 

\begin{figure*}[t!]
 \centering
\noindent \scalebox{0.53}{
  \includegraphics[width=1\linewidth]{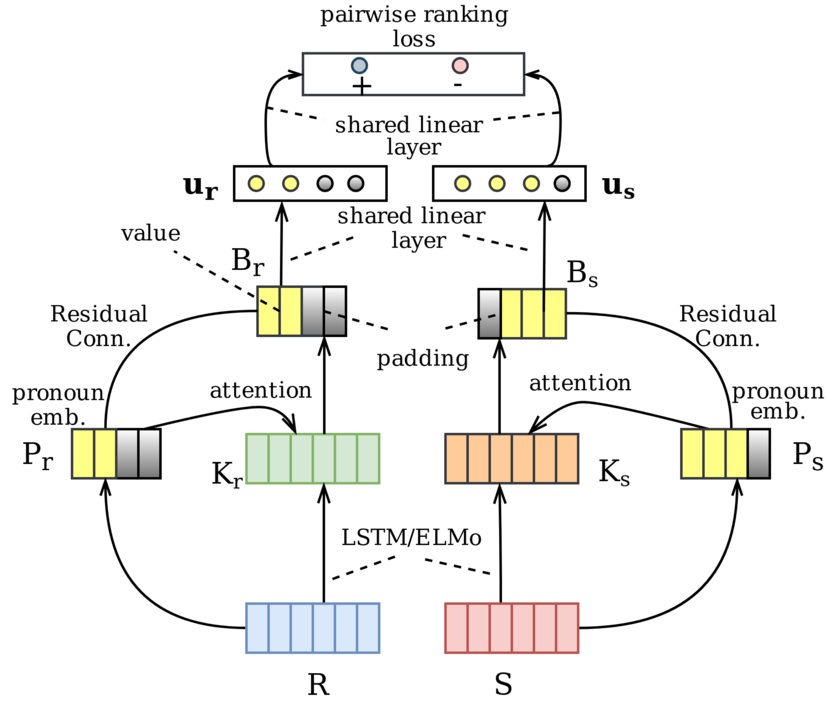}}
\caption{Our proposed framework to differentiate good pronoun translations from bad ones in context.}
\label{fig:proposed-model}

\end{figure*}

Let $R = (C_r;r)$ and $S = (C_s; s)$ denote a \emph{reference} and a \emph{system} tuple pair containing a reference and a system translation, $r$ and $s$, along with a context of previous sentences, $C_r$ and $C_s$, respectively. Note that $C_r$ and $C_s$ can contain the same sentences or different sentences, or be empty in case no context is provided. Given a training set $\Ds = \{(R_i, S_i)\}_{i=1}^N$ containing $N$ such tuple pairs, our aim is to learn an evaluation measure that can rank any unseen translation pair $(R, S)$ with respect to the correct use of pronouns.\footnote{Here $R$ (or $S$) can be a reference or a system translation.} In Section \ref{sec:dataset}, we described how such datasets can be collected opportunistically without recourse to expensive manual annotation.

Figure \ref{fig:proposed-model} shows our proposed framework to evaluate MT outputs with respect to pronouns. The inputs to the model are sentences (with or without context $C_r$ and $C_s$): $R$ and $S$. Each input sentence is first mapped into a set of word embedding vectors of dimensions $d$ by performing a lookup in the shared embedding matrix $E \in \real^{v \times d}$ with vocabulary size $v$. $E$ can be initialized randomly or with any pre-trained embeddings such as GloVe \cite{pennington-socher-manning:2014:EMNLP2014}, or contextualized word vectors such as ELMo \cite{Peters:2018}. 

In case of initialization with GloVe vectors, we use a BiLSTM \cite{hochreiter1997long} {layer to get a representation of the words that is encoded with contextual information.} Let $X = (\mathbf{x}_1,\mathbf{x}_2,\dots, \mathbf{x}_n)$  denote an input sequence, where $\mathbf{x}_t$ is the $t^{th}$ word embedding vector of the sequence. The LSTM recurrent layer computes a compositional representation ${\mathbf{k}}_t$ at every time step $t$ by performing nonlinear transformations of the current input $\mathbf{x}_t$ and the output of the previous time step ${\mathbf{k}}_{t-1}$. In a BiLSTM, we get the representation $\overrightarrow{\mathbf{k}_t}$ by processing the sequence in the forward direction, and the representation $\overleftarrow{\mathbf{k}_t}$ by processing the sequence in the backward direction. The final representation $\mathbf{k}_t$ of a word is the concatenation of these two representations, i.e., $\mathbf{k}_t = [\overrightarrow{\mathbf{k}_t}; \overleftarrow{\mathbf{k}_t}]$.

With ELMo initialization, the word vectors obtained are used directly. {ELMo uses stacked biLSTM encoder and gives very powerful contextualized word representations learned from a large corpora by optimizing a bi-directional language modeling loss.} The ELMo representations already capture morphological, syntactic and contextual semantic features \cite{peters-etal-2018-dissecting}.\footnote{{We also tried an ELMo-initialized BiLSTM, but it did not perform well while increasing model complexity.}} 

Let $K_r$ and $K_s$ be the matrices whose rows represent the word representations of $R$ and $S$, respectively (obtained either from Bi-LSTM or directly from ELMo). From these representations, we extract the representations of the pronouns in the target sentence (from $r$ and $s$; not from the contexts). Let $P_r$ and $P_s$ be the matrices whose rows represent the contextualized word representations of the pronouns in $r$ and $s$, respectively. We use zero-padding (shown as {shaded} boxes) to make $P_r$ and $P_s$ fixed-length. 

We then use \emph{scaled multiplicative attention} \cite{Vaswani-17-transformer} to compute a contextual representation for the pronouns in $r$ and $s$. Specifically, we consider the rows of $P_r$ (\resp\ $P_s$) as \emph{query} vectors, the rows of $K_r$ (\resp\ $K_s$) as \emph{key} and \emph{value} vectors, and the matrix $P_r$ (\resp\ $P_s$) to attend over $K_r$ (\resp\ $K_s$). We use residual connection and layer normalization to get the pronoun representations $B_r$ and $B_s$:

\begin{align}
 B'_r = \Ss(\frac{P_r K_r^T}{\sqrt[2]{d}})K_r; \hspace{0.5em} B'_s = \Ss(\frac{P_s K_s^T}{\sqrt[2]{d}})K_s
\end{align}
\vspace{-1em}
\begin{align}
\hspace{-0.4em}B_r \hspace{-0.2em}= \text{LayerNor}(P_r \hspace{-0.2em} + \hspace{-0.2em}B'_r); \hspace{0.2em} B_s = \text{LayerNor}(P_s\hspace{-0.2em}+\hspace{-0.2em}B'_s) 
 \label{eq:attn}
\end{align}

\normalsize

\noindent Note that $B_r$ and $B_s$ contain a $d$-dimensional vector for each query (pronoun) vector (and \textit{zero} vectors due to padding). We pass these vectors through a \textbf{shared} linear layer parameterized by $\mathbf{z} \in \mathbb{R}^d$ to obtain a score for each pronoun. This yields vectors $\mathbf{u}_r$ and $\mathbf{u}_s$ for the reference and for the system translations:     


\begin{align}
\mathbf{u}_r = B_r \mathbf{z}; \hspace{1em} \mathbf{u}_s = B_s \mathbf{z};
\end{align}

\normalsize

A final \textbf{shared} linear layer parameterized by $\mathbf{w}$ converts these vectors to contrastive scores, yielding a (positive) score $y_r$ for the reference and a (negative) score $y_s$ for the system translation: 

\begin{align}
y_r = \mathbf{u}_r^T\mathbf{w}; \hspace{1em} 
y_s = \mathbf{u}_s^T\mathbf{w};
\end{align}
\normalsize

We then use the scores in a pairwise ranking loss \cite{collobert2011natural} to find model parameters that assign a higher score to $y_r$ than to $y_s$. We minimize the 
following ranking objective:
 
\vspace{-1em}
\begin{align}
\Ls(\theta)=  \max \{0, 0.1 - y_r + y_s\} 
\end{align}
\normalsize

\noindent Note that the network shares all of its parameters ($\theta$) to obtain $y_r$ and $y_s$ from a pair of {inputs $R_i = (C_r, r)$ and $S_i = (C_s, s)$}. Once trained, it can be used to score any input independently. 

\section{Experiments}
\label{sec:exp}

Below, we describe our data, the experimental setup, and the evaluation results. 

\subsection{Data}

We first created a set of commonly confused pronoun pairs. Using the data from the study, we calculated the inter-annotator agreement for each pair of a reference/correct pronoun and a system translation/incorrect pronoun. We excluded the pairs with low agreement (\textless 0.8) or for which the system output was chosen as the correct translation more often. Pairs with low agreement are essentially cases where the annotators cannot agree that the reference translation is better. The source of ambiguity in this case is that the system translation is not absolutely wrong (see Figure  \ref{fig:filtered_example}); therefore, these cases may not be so critical to correct. The remaining pairs are, with a fairly high confidence, positive--negative (correct--incorrect) pairs.

Next, we filtered the WMT data, only keeping sentences with these pronoun pairs. This yielded 97,461 reference translation (positive text) --- unique system output (negative text) pairs for training, taken from WMT11,12,13,15 (Table~\ref{table:data_stats}). 

The development data collected from WMT14 system outputs has 5,727 unique system translations and 6,635 unique noisy candidate pairs.\footnote{The number of unique noisy candidates exceeds that of unique system translations because a separate noisy candidate was generated for each error in a system translation.}

For testing, we used the annotated data from the user study, generated from a subset of WMT17 system translations (except French, which is from the discussion forum test set from WMT15, not overlapping with the training data). There are 500 unique noisy-reference pairs per source language, a total of 2,000.

\begin{table}[t!]
\centering
\scalebox{0.85}{\begin{tabular}{l|l|c}
\toprule
\textbf{Data} & \textbf{Source} & \textbf{\#Unique pairs} \\

\midrule
Training & WMT11-13,15 & 97,461 \\
Development & WMT14 & 5,727 \\
Development (noisy) & WMT14 & 6,635 \\
Test (noisy) & WMT15,17 & 2,000 \\
\bottomrule
\end{tabular}}
\caption{Statistics about our dataset.}
\label{table:data_stats}
\end{table}

\subsection{Experimental Setup}

We evaluated the models in terms of accuracy, \ie\ proportion of times the model scored the reference translation higher than the system/noisy output, and we report results using either GloVe  \cite{pennington-socher-manning:2014:EMNLP2014} or ELMo \cite{Peters:2018}. 
We conducted a number of experiments, training and testing under different conditions:

\paragraph{No Context (NC).} The reference ($R$) (or noisy reference $R'$) and the system ($S$) translations go through ELMo or BiLSTM, without contextual information $C_r/C_s$ (i.e., query = $P_r/P_s$, key, value = $R=r$/$S=s$);

\paragraph{With Context: } The reference ($R$) (or noisy reference $R'$) and system ($S$) translation representations include two previous sentences as context. The context can be further categorized as: 

    \begin{enumerate}[leftmargin=*,noitemsep,topsep=1pt,label=(\roman*)]
    
     \item \textbf{Respective Context (RC)}: $R$ includes its own reference context $C_r = r_{-2}r_{-1}$, and $S$ includes its own system context $C_s = s_{-2}s_{-1}$ (query = $P_r/P_s$, key, value = $R=r_{-2}r_{-1}r$/$S=s_{-2}s_{-1}s$);
      
    \item \textbf{Common Reference Context (CRC)}: The context for $R$ and $S$ includes the same reference context $C_r = C_s = r_{-2}r_{-1}$  (query = $P_r/P_s$, key, value = $R=r_{-2}r_{-1}r$/$S=r_{-2}r_{-1}s$);

    \end{enumerate}

\noindent We perform the evaluation in two ways:
\begin{enumerate}[noitemsep,topsep=1pt,label=(\alph*)]
  \item {$\mathbf{R}$ vs. $\mathbf{S}$:} Testing over pairs of reference ($R$) and system translation ($S$) texts;

  \item {$\mathbf{R}$ vs. $\mathbf{R'}$:} Testing over pairs of reference ($R$) and noisy candidate ($R'$) texts.
  \end{enumerate}
  
\paragraph{Baseline.} For a baseline performance, we simply take the average of the extracted pronoun representations in $P_r$ and $P_s$, and convert them to pairwise scores through linear layers. The baseline is also evaluated with and without context. 
  
\begin{table}[t!]
\centering
\scalebox{0.80}{\begin{tabular}{c|l|l|c|c}
\toprule
 & \textbf{Context} &  & \textbf{Acc.} &\textbf{Acc.} \\
\textbf{Exp} & \textbf{Setting} & \textbf{Test} & \textbf{(Glove)} &\textbf{(ELMo)} \\

\midrule

1 & NC-Baseline & $\mathbf{R}$ vs. $\mathbf{R'}$ & 69.12 & 85.80 \\
2 & NC & $\mathbf{R}$ vs. $\mathbf{R'}$ & 68.97 & 88.04 \\
3 & NC & $\mathbf{R}$ vs. $\mathbf{S }$ & 79.67 & 89.09 \\
\midrule

4 & RC-Baseline & $\mathbf{R}$ vs. $\mathbf{R'}$ & 69.07 & 85.80 \\
5 & RC & $\mathbf{R}$ vs. $\mathbf{R'}$ & 67.88 & 87.90 \\

\midrule

6 & CRC-Baseline & $\mathbf{R}$ vs. $\mathbf{R'}$ & 69.16 & 86.66  \\
7 & CRC & $\mathbf{R}$ vs. $\mathbf{R'}$ & 68.93 & 89.11  \\
8 & CRC & $\mathbf{R}$ vs. $\mathbf{S }$ & 77.87 & 90.69 \\

\bottomrule
\end{tabular}}

\caption{Results on WMT14 with different contexts.}
\label{table:results}

\end{table}

\begin{table}[t!]
\centering
\scalebox{0.80}{\begin{tabular}{r|llc}
\toprule

\textbf{Language} & \textbf{Acc.(ELMo)}   & \textbf{AC1  Agr.}  \\

\midrule

Russian$\rightarrow$English & 79.4  & 0.80 \\

French$\rightarrow$English  & 82.0 & 0.84 \\

German$\rightarrow$English & 81.6  & 0.83\\

Chinese$\rightarrow$English  & 82.4  &  0.83\\ 

\quad- - Only English & ----  & 0.83 \\ 
\midrule
\bf Overall (average) & 81.35  & ---- \\
\bottomrule
\end{tabular}}

\caption{\textbf{Acc}uracy and AC1 \textbf{Agr}eement for the ELMo-based model predictions on the study dataset.}
\label{table:results-study}

\end{table}

\begin{figure*}[tbh]
\centering

{\includegraphics[scale=0.35]{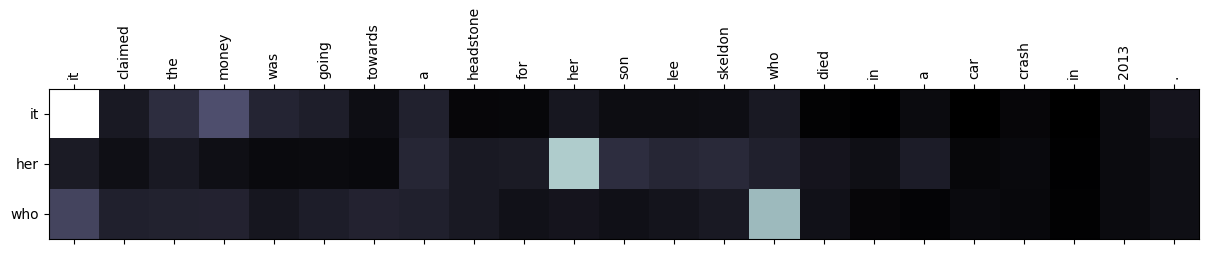}}

{\includegraphics[scale=0.35]{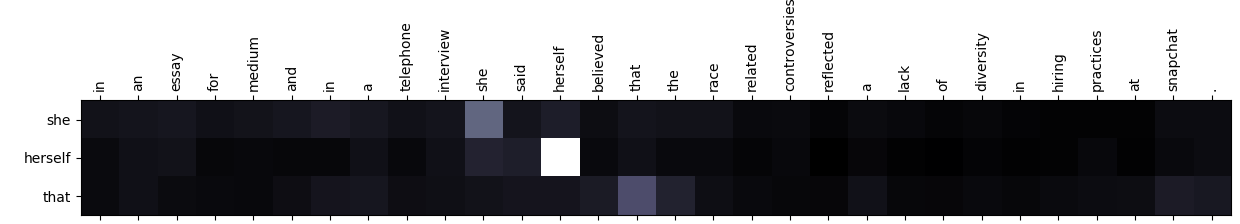}}

\caption{Attention maps for some noisy candidates from the test set.}\label{fig:attn_maps}

\end{figure*}

\begin{figure*}[tbh]
\centering

{\includegraphics[scale=0.35]{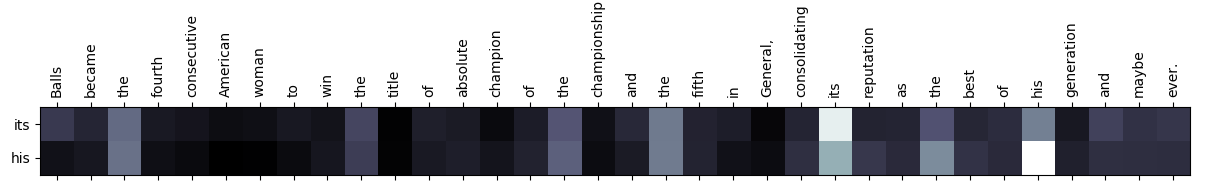}}

{\includegraphics[scale=0.35]{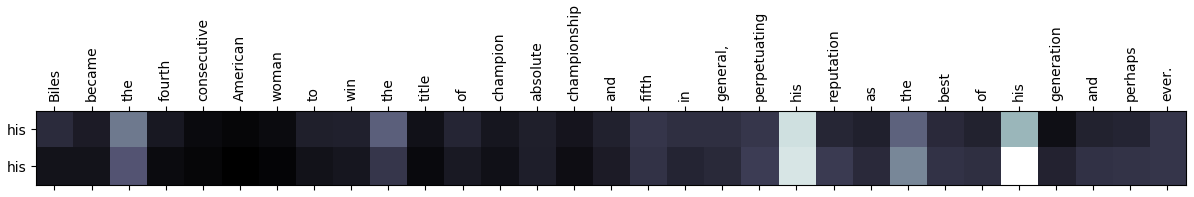}}

\caption{Attention maps for two candidate system translations of the same text (best viewed in color)}\label{fig:attn_maps_sys}

\end{figure*}

\subsection{Results and Analysis}

The first three experiments in Table~\ref{table:results} show results for the `No Context' setting. The results on the noisy data (Exp. 1-2) are indicative of how sensitive our model is to pronouns, since the difference between $R$ and $R'$ is in a single pronoun. The results on the reference/system translation pair $R$ and $S$ (Exp. 3) is indicative of performance in a real use-case. In case of the GloVe-BiLSTM model, using attention instead of averaging pronouns does not help. However, using ELMo greatly improves the accuracy of the baseline and also leads to an improvement while using attention. ELMo can model syntactic and semantic information and improve co-reference resolution \cite{Peters:2018}, which could be a contributing factor.

The second part of our experiments (Exp. 4-5) concerns the addition of contextual information. When the respective contexts are added, there is no improvement in the model (even a drop in case of the GloVe-BiLSTM model); quite possibly, other differences in the text make it harder for the model to focus on the pronoun errors.

To offset this issue, we use a common context for both reference and system sentences, taken from the reference (Exp. 6-8). Training the model with a common reference context (CRC) leads only to marginal difference in the GloVe-BiLSTM model, but the ELMo model improves with the addition of context. All our experiments\footnote{We also ran experiments with BERT \cite{BERT} and found that it performed on par with ELMo.} have shown that ELMo can be quite powerful at capturing contextual information, even as the context size grows.

Finally, we test our \emph{ELMo - common reference context} model on the held-out dataset used for the user study. Table~\ref{table:results-study} shows the results. The accuracy of the system is lower on the study dataset. Note that since the training data was filtered based on the pronoun pairs with high agreement from the study, the study dataset contains pronoun pairs with low agreement that were not seen during training. We also calculate the agreement with human judgments excluding ties since our model does not handle ties. The overall agreement with the human judgments remains high.

\subsection{Pronoun-wise Analysis}
We performed a pronoun-wise analysis of the results in Table~\ref{table:results-study}. The model scored the noisy version higher than the reference in about 19\% of the cases. Of these, 46\% were pronoun pairs that were not seen during training.  

Of the remaining pronoun pairs that were seen during training, the main source of errors (over 11\%) were cases when \emph{\textbf{that}} in the reference was replaced by \emph{\textbf{it}} in the noisy version (\emph{``We always tell victims not to pay up; \textbf{that/it} simply exacerbates the problem", explains Kleczynski}). The noisy candidate was scored higher about 28\% of the time, out of 79 samples. The next highest source of errors was the reference--noisy pair of \emph{\textbf{it--she}} at 10\%.

In contrast, the best-performing pair was when a \emph{\textbf{he}} in the reference was replaced with an \emph{\textbf{it}} in the noisy version (\emph{\textbf{He/It} risked everything to save other people's lives.}). The reference was scored higher than the noisy candidate 95\% of the time, out of 135 samples. The next highest performer was the reference--noisy pair of \emph{\textbf{his--its}}, which was correctly scored 86\% of the time.

This performance follows the distribution of the pronoun pairs seen during training. The \emph{\textbf{he--it}} and \emph{\textbf{his--its}} pairs together account for over 12\% of the training data, while the \emph{\textbf{that--it}} and \emph{\textbf{it--she}} pairs together form only 3.7\%. Since the distribution of the pronoun pairs in the training data is itself based on the distribution of errors in the system translations, the model performs best over error cases that occur most often in system translations.

\subsection{Discussion}
\label{subsec:further_analysis}

As systems participating in WMT improve over the years, our test data is closer to state-of-the-art neural MT, while most of the training data is from statistical systems. This setting allows us to capture a wider range of errors while showing that our model is sensitive to small errors in a fluent output. 

Note that since the model is trained on the full system output and also on all pronouns in that output, it has not received any signal about which pronoun is wrong. Yet, we can see from the attention maps in Figure~\ref{fig:attn_maps} that the model can correctly identify the incorrect pronoun. In Figure~\ref{fig:attn_maps} (top), the model distinguishes the wrong pronoun \textit{it} (correct: \textit{she}), while in Figure~\ref{fig:attn_maps} (bottom) it correctly finds \textit{herself} as the wrong translation (correct: \textit{she}).

We further compare the scores of two system translations for the same sentence from WMT17 Russian-English system outputs (see Figure~\ref{fig:attn_maps_sys}). The correct pronoun is \textit{her}. While one system translates it alternately as \textit{its} and \textit{his} (Figure \ref{fig:attn_maps_sys}, top), the other system translates both cases as \textit{his} (Figure~\ref{fig:attn_maps_sys}, bottom). Our model scores the translation in Figure~\ref{fig:attn_maps_sys} (bottom) higher than the translation in Figure~\ref{fig:attn_maps_sys} (top); though it highlights both occurrences of \textit{his} as wrong, it believes that \textit{its} is worse. It could be argued that the translation in Figure~\ref{fig:attn_maps_sys} (bottom) is better since it maintains the animacy/human aspect, even if the gender is wrong, and it is also consistent. Evaluation measures such as APT and AutoPRF are likely to yield the same accuracy/precision-recall for both cases.

\section{Conclusions and Future Work}
\label{sec:con}
We have presented a new, extensive, targeted dataset for pronoun translation that covers multiple source languages and a wide range of target English pronouns. We also proposed a novel evaluation measure for differentiating good vs. bad pronoun translations, irrespective of the source language, which achieved high correlation with human judgments. 

In future work, we want to handle cases where multiple pronouns are equally suitable in a given context. We would also like to extend the work to other discourse phenomena.

\section*{Acknowledgments}

We would like to thank the anonymous reviewers for their comments. Shafiq Joty would like to thank the funding support from his Start-up Grant (M4082038.020). 

\bibliographystyle{acl_natbib}
\bibliography{pronoun_refs}

\begin{thebibliography}{44}
\expandafter\ifx\csname natexlab\endcsname\relax\def\natexlab#1{#1}\fi

\bibitem[{Bahdanau et~al.(2014)Bahdanau, Cho, and
  Bengio}]{DBLP:journals/corr/BahdanauCB14}
Dzmitry Bahdanau, Kyunghyun Cho, and Yoshua Bengio. 2014.
\newblock Neural machine translation by jointly learning to align and
  translate.
\newblock \emph{CoRR}, abs/1409.0473.

\bibitem[{Bawden et~al.(2018)Bawden, Sennrich, Birch, and
  Haddow}]{Bawden-NAACL18}
Rachel Bawden, Rico Sennrich, Alexandra Birch, and Barry Haddow. 2018.
\newblock \href {http://aclweb.org/anthology/N18-1118} {{Evaluating Discourse
  Phenomena in Neural Machine Translation}}.
\newblock In \emph{{Proceedings of the 2018 Conference of the North American
  Chapter of the Association for Computational Linguistics: Human Language
  Technologies, Volume 1 (Long Papers)}}, pages 1304--1313, New Orleans,
  Louisiana. Association for Computational Linguistics.

\bibitem[{Bojar et~al.(2014)Bojar, Buck, Federmann, Haddow, Koehn, Leveling,
  Monz, Pecina, Post, Saint-Amand, Soricut, Specia, and Tamchyna}]{WMT2014}
Ondrej Bojar, Christian Buck, Christian Federmann, Barry Haddow, Philipp Koehn,
  Johannes Leveling, Christof Monz, Pavel Pecina, Matt Post, Herve Saint-Amand,
  Radu Soricut, Lucia Specia, and Ale\v{s} Tamchyna. 2014.
\newblock \href {http://www.aclweb.org/anthology/W/W14/W14-3302} {Findings of
  the 2014 workshop on statistical machine translation}.
\newblock In \emph{Proceedings of the Ninth Workshop on Statistical Machine
  Translation}, pages 12--58, Baltimore, Maryland, USA. Association for
  Computational Linguistics.

\bibitem[{Bojar et~al.(2013)Bojar, Buck, Callison-Burch, Federmann, Haddow,
  Koehn, Monz, Post, Soricut, and Specia}]{WMT2013}
Ond\v{r}ej Bojar, Christian Buck, Chris Callison-Burch, Christian Federmann,
  Barry Haddow, Philipp Koehn, Christof Monz, Matt Post, Radu Soricut, and
  Lucia Specia. 2013.
\newblock \href {http://www.aclweb.org/anthology/W13-2201} {Findings of the
  2013 {Workshop on Statistical Machine Translation}}.
\newblock In \emph{Proceedings of the Eighth Workshop on Statistical Machine
  Translation}, pages 1--44, Sofia, Bulgaria. Association for Computational
  Linguistics.

\bibitem[{Bojar et~al.(2017)Bojar, Chatterjee, Federmann, Graham, Haddow,
  Huang, Huck, Koehn, Liu, Logacheva, Monz, Negri, Post, Rubino, Specia, and
  Turchi}]{WMT2017}
Ond\v{r}ej Bojar, Rajen Chatterjee, Christian Federmann, Yvette Graham, Barry
  Haddow, Shujian Huang, Matthias Huck, Philipp Koehn, Qun Liu, Varvara
  Logacheva, Christof Monz, Matteo Negri, Matt Post, Raphael Rubino, Lucia
  Specia, and Marco Turchi. 2017.
\newblock \href {http://www.aclweb.org/anthology/W17-4717} {Findings of the
  2017 conference on machine translation (wmt17)}.
\newblock In \emph{Proceedings of the Second Conference on Machine Translation,
  Volume 2: Shared Task Papers}, pages 169--214, Copenhagen, Denmark.
  Association for Computational Linguistics.

\bibitem[{Bojar et~al.(2015)Bojar, Chatterjee, Federmann, Haddow, Huck, Hokamp,
  Koehn, Logacheva, Monz, Negri, Post, Scarton, Specia, and Turchi}]{WMT2015}
Ond\v{r}ej Bojar, Rajen Chatterjee, Christian Federmann, Barry Haddow, Matthias
  Huck, Chris Hokamp, Philipp Koehn, Varvara Logacheva, Christof Monz, Matteo
  Negri, Matt Post, Carolina Scarton, Lucia Specia, and Marco Turchi. 2015.
\newblock \href {http://aclweb.org/anthology/W15-3001} {Findings of the 2015
  workshop on statistical machine translation}.
\newblock In \emph{Proceedings of the Tenth Workshop on Statistical Machine
  Translation}, pages 1--46, Lisbon, Portugal. Association for Computational
  Linguistics.

\bibitem[{Brown et~al.(1993)Brown, Pietra, Pietra, and Mercer}]{J93-2003}
Peter~E. Brown, Stephen A.~Della Pietra, Vincent J.~Della Pietra, and Robert~L.
  Mercer. 1993.
\newblock The mathematics of statistical machine translation: Parameter
  estimation.
\newblock \emph{Computational Linguistics}, 19(2).

\bibitem[{Callison-Burch et~al.(2012)Callison-Burch, Koehn, Monz, Post,
  Soricut, and Specia}]{WMT2012}
Chris Callison-Burch, Philipp Koehn, Christof Monz, Matt Post, Radu Soricut,
  and Lucia Specia. 2012.
\newblock \href {http://www.aclweb.org/anthology/W12-3102} {Findings of the
  2012 workshop on statistical machine translation}.
\newblock In \emph{Proceedings of the Seventh Workshop on Statistical Machine
  Translation}, pages 10--51, Montr{\'e}al, Canada. Association for
  Computational Linguistics.

\bibitem[{Callison-Burch et~al.(2011)Callison-Burch, Koehn, Monz, and
  Zaidan}]{WMT2011}
Chris Callison-Burch, Philipp Koehn, Christof Monz, and Omar Zaidan. 2011.
\newblock \href {http://www.aclweb.org/anthology/W11-2103} {Findings of the
  2011 workshop on statistical machine translation}.
\newblock In \emph{Proceedings of the Sixth Workshop on Statistical Machine
  Translation}, pages 22--64, Edinburgh, Scotland. Association for
  Computational Linguistics.

\bibitem[{Carpuat and Wu(2007)}]{D07-1007}
Marine Carpuat and Dekai Wu. 2007.
\newblock Improving statistical machine translation using word sense
  disambiguation.
\newblock In \emph{Proceedings of the 2007 Joint Conference on Empirical
  Methods in Natural Language Processing and Computational Natural Language
  Learning}, EMNLP-CoNLL~'07.

\bibitem[{Chan et~al.(2007)Chan, Ng, and Chiang}]{P07-1005}
Yee~Seng Chan, Hwee~Tou Ng, and David Chiang. 2007.
\newblock Word sense disambiguation improves statistical machine translation.
\newblock In \emph{Proceedings of the 45th Annual Meeting of the Association of
  Computational Linguistics}, ACL~'07, pages 33--40, Prague, Czech Republic.

\bibitem[{Chiang(2005)}]{Chiang:2005:HPM:1219840.1219873}
David Chiang. 2005.
\newblock A hierarchical phrase-based model for statistical machine
  translation.
\newblock In \emph{Proceedings of the 43rd Annual Meeting on Association for
  Computational Linguistics}, ACL '05, pages 263--270, Ann Arbor, Michigan.

\bibitem[{Collobert et~al.(2011)Collobert, Weston, Bottou, Karlen, Kavukcuoglu,
  and Kuksa}]{collobert2011natural}
Ronan Collobert, Jason Weston, L{\'e}on Bottou, Michael Karlen, Koray
  Kavukcuoglu, and Pavel Kuksa. 2011.
\newblock Natural language processing (almost) from scratch.
\newblock \emph{The Journal of Machine Learning Research}, 12:2493--2537.

\bibitem[{Devlin et~al.(2018)Devlin, Chang, Lee, and Toutanova}]{BERT}
Jacob Devlin, Ming-Wei Chang, Kenton Lee, and Kristina Toutanova. 2018.
\newblock Bert: Pre-training of deep bidirectional transformers for language
  understanding.
\newblock In \emph{NAACL-HLT}.

\bibitem[{Duh(2008)}]{Duh2008}
Kevin Duh. 2008.
\newblock \href {https://doi.org/10.3115/1626394.1626425} {{Ranking vs.
  regression in machine translation evaluation}}.
\newblock \emph{Proceedings of the Third Workshop on Statistical Machine
  Translation}, (June):191--194.

\bibitem[{Dyer et~al.(2013)Dyer, Chahuneau, and Smith}]{dyer2013fastalign}
Chris Dyer, Victor Chahuneau, and Noah~A Smith. 2013.
\newblock A simple, fast, and effective reparameterization of ibm model 2.
\newblock In \emph{Proceedings of the 2013 Conference of the North American
  Chapter of the Association for Computational Linguistics: Human Language
  Technologies}, pages 644--648.

\bibitem[{Galley et~al.(2004)Galley, Hopkins, Knight, and
  Marcu}]{galley-EtAl:2004:HLTNAACL}
Michel Galley, Mark Hopkins, Kevin Knight, and Daniel Marcu. 2004.
\newblock What's in a translation rule?
\newblock In \emph{HLT-NAACL 2004: Main Proceedings}, pages 273--280, Boston,
  Massachusetts, USA.

\bibitem[{Guillou and Hardmeier(2016)}]{GUILLOU16.327}
Liane Guillou and Christian Hardmeier. 2016.
\newblock Protest: A test suite for evaluating pronouns in machine translation.
\newblock In \emph{Proceedings of the Tenth International Conference on
  Language Resources and Evaluation (LREC 2016)}, Paris, France. European
  Language Resources Association (ELRA).

\bibitem[{Guillou and Hardmeier(2018)}]{Liane-EMNLP18}
Liane Guillou and Christian Hardmeier. 2018.
\newblock Automatic reference-based evaluation of pronoun translation misses
  the point.
\newblock In \emph{Proceedings of the 2018 Conference on Empirical Methods in
  Natural Language Processing}, EMNLP~'18, pages 4797--4802, Brussels, Belgium.

\bibitem[{Guillou et~al.(2018)Guillou, Hardmeier, Lapshinova-Koltunski, and
  Lo{\'a}iciga}]{guillou2018pronoun}
Liane Guillou, Christian Hardmeier, Ekaterina Lapshinova-Koltunski, and Sharid
  Lo{\'a}iciga. 2018.
\newblock A pronoun test suite evaluation of the english--german mt systems at
  wmt 2018.
\newblock In \emph{Proceedings of the Third Conference on Machine Translation:
  Shared Task Papers}, pages 570--577.

\bibitem[{Guillou et~al.(2016)Guillou, Hardmeier, Nakov, Stymne, Tiedemann,
  Versley, Cettolo, Webber, and Popescu-Belis}]{Guillou-16-WMT}
Liane Guillou, Christian Hardmeier, Preslav Nakov, Sara Stymne, J{\"o}rg
  Tiedemann, Yannick Versley, Mauro Cettolo, Bonnie Webber, and Andrei
  Popescu-Belis. 2016.
\newblock \href {https://doi.org/10.18653/v1/W16-2345} {Findings of the 2016
  wmt shared task on cross-lingual pronoun prediction}.
\newblock In \emph{Proceedings of the First Conference on Machine Translation:
  Volume 2, Shared Task Papers}, pages 525--542. Association for Computational
  Linguistics.

\bibitem[{Guzm{\'{a}}n et~al.(2015)Guzm{\'{a}}n, Joty, M{\`{a}}rquez, and
  Nakov}]{Guzman2015a}
Francisco Guzm{\'{a}}n, Shafiq Joty, Llu{\'{i}}s M{\`{a}}rquez, and Preslav
  Nakov. 2015.
\newblock \href {https://doi.org/10.3115/v1/p15-1078} {{Pairwise Neural Machine
  Translation Evaluation}}.
\newblock \emph{Proceedings of the 53rd Annual Meeting of the Association for
  Computational Linguistics and the 7th International Joint Conference on
  Natural Language Processing}, pages 805--814.

\bibitem[{Guzm{\'a}n et~al.(2017)Guzm{\'a}n, Joty, i~Villodre, and
  Nakov}]{Guzmn2017MachineTE}
Francisco Guzm{\'a}n, Shafiq~R. Joty, Llu{\'i}s~M{\`a}rquez i~Villodre, and
  Preslav Nakov. 2017.
\newblock Machine translation evaluation with neural networks.
\newblock \emph{Computer Speech \& Language}, 45:180--200.

\bibitem[{Gwet(2008)}]{gwetac1}
Kilem~Li Gwet. 2008.
\newblock Computing inter-rater reliability and its variance in the presence of
  high agreement.
\newblock \emph{British Journal of Mathematical and Statistical Psychology},
  61(1):29--48.

\bibitem[{Hardmeier(2014)}]{Hardmeier-14}
Christian Hardmeier. 2014.
\newblock \emph{Discourse in Statistical Machine Translation}.
\newblock Ph.D. thesis, Uppsala University, Department of Linguistics and
  Philology, Uppsala, Sweden.

\bibitem[{Hardmeier and Federico(2010)}]{Hardmeier2010ModellingPA}
Christian Hardmeier and Marcello Federico. 2010.
\newblock Modelling pronominal anaphora in statistical machine translation.
\newblock In \emph{IWSLT}.

\bibitem[{Hardmeier et~al.(2015)Hardmeier, Nakov, Stymne, Tiedemann, Versley,
  and Cettolo}]{W15-2501}
Christian Hardmeier, Preslav Nakov, Sara Stymne, J{\"o}rg Tiedemann, Yannick
  Versley, and Mauro Cettolo. 2015.
\newblock Pronoun-focused mt and cross-lingual pronoun prediction: Findings of
  the 2015 discomt shared task on pronoun translation.
\newblock In \emph{Proceedings of the Second Workshop on Discourse in Machine
  Translation}, pages 1--16, Lisbon, Portugal.

\bibitem[{Hardmeier et~al.(2012)Hardmeier, Nivre, and
  Tiedemann}]{Hardmeier:2012:DDP:2390948.2391081}
Christian Hardmeier, Joakim Nivre, and J\"{o}rg Tiedemann. 2012.
\newblock Document-wide decoding for phrase-based statistical machine
  translation.
\newblock In \emph{Proceedings of the 2012 Joint Conference on Empirical
  Methods in Natural Language Processing and Computational Natural Language
  Learning}, EMNLP-CoNLL '12, pages 1179--1190, Jeju Island, Korea.

\bibitem[{Hassan et~al.(2018)Hassan, Aue, Chen, Chowdhary, Clark, Federmann,
  Huang, Junczys{-}Dowmunt, Lewis, Li, Liu, Liu, Luo, Menezes, Qin, Seide, Tan,
  Tian, Wu, Wu, Xia, Zhang, Zhang, and
  Zhou}]{DBLP:journals/corr/abs-1803-05567}
Hany Hassan, Anthony Aue, Chang Chen, Vishal Chowdhary, Jonathan Clark,
  Christian Federmann, Xuedong Huang, Marcin Junczys{-}Dowmunt, William Lewis,
  Mu~Li, Shujie Liu, Tie{-}Yan Liu, Renqian Luo, Arul Menezes, Tao Qin, Frank
  Seide, Xu~Tan, Fei Tian, Lijun Wu, Shuangzhi Wu, Yingce Xia, Dongdong Zhang,
  Zhirui Zhang, and Ming Zhou. 2018.
\newblock Achieving human parity on automatic chinese to english news
  translation.
\newblock \emph{CoRR}, abs/1803.05567.

\bibitem[{Hochreiter and Schmidhuber(1997)}]{hochreiter1997long}
Sepp Hochreiter and J\"{u}rgen Schmidhuber. 1997.
\newblock Long short-term memory.
\newblock \emph{Neural Computation}, 9(8):1735--1780.

\bibitem[{Koehn et~al.(2003)Koehn, Och, and
  Marcu}]{Koehn:2003:SPT:1073445.1073462}
Philipp Koehn, Franz~Josef Och, and Daniel Marcu. 2003.
\newblock Statistical phrase-based translation.
\newblock In \emph{Proceedings of the 2003 Conference of the North American
  Chapter of the Association for Computational Linguistics on Human Language
  Technology - Volume 1}, NAACL '03, pages 48--54, Edmonton, Canada.

\bibitem[{L{\"a}ubli et~al.(2018)L{\"a}ubli, Sennrich, and
  Volk}]{Samuel-EMNLP-18}
Samuel L{\"a}ubli, Rico Sennrich, and Martin Volk. 2018.
\newblock \href {http://aclweb.org/anthology/D18-1512} {{Has Machine
  Translation Achieved Human Parity? A Case for Document-level Evaluation}}.
\newblock In \emph{{Proceedings of the 2018 Conference on Empirical Methods in
  Natural Language Processing}}, pages 4791--4796. Association for
  Computational Linguistics.

\bibitem[{Lo{\'a}iciga et~al.(2017)Lo{\'a}iciga, Stymne, Nakov, Hardmeier,
  Tiedemann, Cettolo, and Versley}]{Sharid-W17}
Sharid Lo{\'a}iciga, Sara Stymne, Preslav Nakov, Christian Hardmeier, J{\"o}rg
  Tiedemann, Mauro Cettolo, and Yannick Versley. 2017.
\newblock \href {https://doi.org/10.18653/v1/W17-4801} {Findings of the 2017
  discomt shared task on cross-lingual pronoun prediction}.
\newblock In \emph{Proceedings of the Third Workshop on Discourse in Machine
  Translation}, pages 1--16. Association for Computational Linguistics.

\bibitem[{Miculicich~Werlen and Popescu-Belis(2017)}]{APT}
Lesly Miculicich~Werlen and Andrei Popescu-Belis. 2017.
\newblock \href {https://doi.org/10.18653/v1/W17-4802} {Validation of an
  automatic metric for the accuracy of pronoun translation (apt)}.
\newblock In \emph{Proceedings of the Third Workshop on Discourse in Machine
  Translation}, pages 17--25. Association for Computational Linguistics.

\bibitem[{M\"{u}ller et~al.(2018)M\"{u}ller, Rios, Voita, and
  Sennrich}]{muller-EtAl:2018:WMT}
Mathias M\"{u}ller, Annette Rios, Elena Voita, and Rico Sennrich. 2018.
\newblock A large-scale test set for the evaluation of context-aware pronoun
  translation in neural machine translation.
\newblock In \emph{Proceedings of the Third Conference on Machine Translation:
  Research Papers}, WMT~'18, pages 61--72, Belgium, Brussels.

\bibitem[{Papineni et~al.(2002)Papineni, Roukos, Ward, and
  Zhu}]{papineni2002bleu}
Kishore Papineni, Salim Roukos, Todd Ward, and Wei-Jing Zhu. 2002.
\newblock Bleu: a method for automatic evaluation of machine translation.
\newblock In \emph{ACL}.

\bibitem[{Pennington et~al.(2014)Pennington, Socher, and
  Manning}]{pennington-socher-manning:2014:EMNLP2014}
Jeffrey Pennington, Richard Socher, and Christopher Manning. 2014.
\newblock \href {http://www.aclweb.org/anthology/D14-1162} {Glove: Global
  vectors for word representation}.
\newblock In \emph{EMNLP'14}, pages 1532--1543, Doha, Qatar.

\bibitem[{Peters et~al.(2018{\natexlab{a}})Peters, Neumann, Zettlemoyer, and
  Yih}]{peters-etal-2018-dissecting}
Matthew Peters, Mark Neumann, Luke Zettlemoyer, and Wen-tau Yih.
  2018{\natexlab{a}}.
\newblock \href {https://www.aclweb.org/anthology/D18-1179} {Dissecting
  contextual word embeddings: Architecture and representation}.
\newblock In \emph{Proceedings of the 2018 Conference on Empirical Methods in
  Natural Language Processing}, pages 1499--1509, Brussels, Belgium.
  Association for Computational Linguistics.

\bibitem[{Peters et~al.(2018{\natexlab{b}})Peters, Neumann, Iyyer, Gardner,
  Clark, Lee, and Zettlemoyer}]{Peters:2018}
Matthew~E. Peters, Mark Neumann, Mohit Iyyer, Matt Gardner, Christopher Clark,
  Kenton Lee, and Luke Zettlemoyer. 2018{\natexlab{b}}.
\newblock Deep contextualized word representations.
\newblock In \emph{NAACL}.

\bibitem[{Radford(2018)}]{GPT-1}
Alec Radford. 2018.
\newblock Improving language understanding by generative pre-training.

\bibitem[{Reiter(2018)}]{reiter2018structured}
Ehud Reiter. 2018.
\newblock A structured review of the validity of {BLEU}.
\newblock \emph{Computational Linguistics}, 44(3):393--401.

\bibitem[{Rios et~al.(2018)Rios, M{\"u}ller, and
  Sennrich}]{rios-mller-sennrich:2018:WMT}
Annette Rios, Mathias M{\"u}ller, and Rico Sennrich. 2018.
\newblock \href {http://www.statmt.org/wmt18/pdf/WMT064.pdf} {{The Word Sense
  Disambiguation Test Suite at WMT18}}.
\newblock In \emph{{Proceedings of the Third Conference on Machine
  Translation}}, pages 594--602, Belgium, Brussels. Association for
  Computational Linguistics.

\bibitem[{Vaswani et~al.(2017)Vaswani, Shazeer, Parmar, Uszkoreit, Jones,
  Gomez, Kaiser, and Polosukhin}]{Vaswani-17-transformer}
Ashish Vaswani, Noam Shazeer, Niki Parmar, Jakob Uszkoreit, Llion Jones,
  Aidan~N Gomez, \L~ukasz Kaiser, and Illia Polosukhin. 2017.
\newblock Attention is all you need.
\newblock In I.~Guyon, U.~V. Luxburg, S.~Bengio, H.~Wallach, R.~Fergus,
  S.~Vishwanathan, and R.~Garnett, editors, \emph{Advances in Neural
  Information Processing Systems 30}, pages 5998--6008. Curran Associates, Inc.

\bibitem[{Voita et~al.(2018)Voita, Serdyukov, Sennrich, and
  Titov}]{Voita-ACL18}
Elena Voita, Pavel Serdyukov, Rico Sennrich, and Ivan Titov. 2018.
\newblock \href {http://aclweb.org/anthology/P18-1117} {{Context-Aware Neural
  Machine Translation Learns Anaphora Resolution}}.
\newblock In \emph{{Proceedings of the 56th Annual Meeting of the Association
  for Computational Linguistics (Volume 1: Long Papers)}}, pages 1264--1274,
  Melbourne, Australia. Association for Computational Linguistics.

\end{thebibliography}




\end{document}